\documentclass[letterpaper, 10pt, conference]{template/ieeeconf}  % Comment this line out if you need a4paper
\usepackage[T1]{fontenc}
\usepackage{wrapfig}
\usepackage{booktabs}
\usepackage{amsmath}
\usepackage{amsmath} % assumes amsmath package installed
\usepackage{amssymb}  % assumes amsmath package installed
\usepackage[]{changes}
\usepackage{accents}
\usepackage{todonotes}
\definechangesauthor[name={Todor}, color=red]{tsv}
\definechangesauthor[name={Johannes}, color=cyan]{jsk}
\presetkeys%
{todonotes}%
{inline,backgroundcolor=yellow}{}
\usepackage{chngcntr}
\usepackage{siunitx}
\usepackage{tikz}
\usepackage{adjustbox}
\usepackage{url}
\usepackage{multirow}
\usepackage{subfigure}
\usepackage[ruled,linesnumbered]{algorithm2e}
\usepackage{xcolor}
\usepackage{pifont}% http://ctan.org/pkg/pifont
\newcommand{\xmark}{\ding{55}}%
\usepackage{makecell}

\DeclareMathOperator*{\E}{\mathbb{E}}

\IEEEoverridecommandlockouts                              % This command is only needed if 
\title{\LARGE \bf
    Transferring Knowledge for Reinforcement Learning in \\Contact-Rich Manipulation
}
\author{Quantao Yang, Johannes A. Stork, and Todor Stoyanov% <-this % stops a space
	\thanks{$^*$This work was supported by the Wallenberg AI, Autonomous Systems and Software Program (WASP) funded by Knut and Alice Wallenberg Foundation.}% <-this % stops a space
	\thanks{Autonomous Mobile Manipulation (AMM) Lab, \"Orebro University, Sweden (e-mail: quantao.yang@oru.se; johannes.stork@oru.se; todor.stoyanov@oru.se).}% <-this % stops a space
}

\begin{document}

	\maketitle
	\section{\uppercase{Introduction}}
	\label{sec:introduction}

	%\todo{Start with a motivation}
	While humans are adept in transferring a learned skill---that is the ability of solving a task---to a new similar task efficiently, most state-of-the-art reinforcement learning (RL) methods have to solve every new task from scratch. %and it is not easy to apply the learned policy to a new problem, even if it is a very similar one~\cite{bousmalis2018using}. 
	Consequently, millions of new interactions with different environments can be required to solve variant tasks, which is infeasible for a real robot system. 
	Training from scratch is resource and time consuming, while sample collection in a new physical environment is costly and repetitive.
	%Meanwhile, RL involves exploratory actions by trial-and-error, which can cause damage to the robot or its environment.
	Therefore, in order to apply RL directly on real physical robots, it is imperative to address the problem of sample-inefficiency when solving variant tasks. 
	
	State-of-the-art methods require policy training in simulation to prevent undesired behavior and later domain transfer, or guided policy search for single skills in a family of similar problems~\cite{peng2018sim, hazara2019transferring}, \cite{du2021auto}. The successful deployment of simulation-to-reality methods requires that the simulation is close enough to the physical system.
	However, for real world robotic applications, transition dynamics in deployment are often substantially different from those encountered during training (in simulation).%, which might lead to the failure of transferring knowledge.
	%Any discrepancies between simulation and reality can cause the robot to perform poorly in the real environment. 
	%The work~\cite{yang2021learning} utilized a framework for learning latent action spaces for RL agents from demonstrated trajectories and connected it to a variable impedance Cartesian space controller, allowing us to learn contact-rich tasks safely and efficiently. However, the method~\cite{yang2021learning} requires demonstration from an expert or expert policy in the precise domain we are solving. 
	
	%Another promising method for these problems is meta learning~\cite{vilalta2002perspective, finn2018probabilistic} that aims at training models which perform well on new tasks or environments. In meta RL methods~\cite{finn2017model, arndt2020meta}, the meta network model should be optimized in advance such that they provide a good starting point for further RL adaptation. We propose Multi-Prior Regularized RL (MPR-RL) by taking advantage of the idea from meta RL to solve the target task that is similar to a family of prior problems.
	
	%\todo{Tell what this paper is about}
	In this work, we consider the problem of transferring knowledge within a family of similar tasks. %\textcolor{red}{We assume that tasks are different only in dynamics and similarity among tasks can be learned by comparing the dynamics.} The prior knowledge can not be transferred from prior tasks to a target task easily. We propose Multi-Prior Regularized RL (MPR-RL) to leverage prior knowledge or skills for a family of problems and transfer the learned knowledge to learn a policy for a new similar problem. 
	Our fundamental assumption is that we are presented with a family of problems, formalized as Markov Decision Processes (MDPs) that all share the same state and action spaces~\cite{yang2022mpr}. Crucially however, we allow for members of the family to exhibit different transition dynamics. Informally, our assumption is that while transition probabilities are different, they may be correlated or overlapping for parts of the state space. We then propose a method---Multi-Prior Regularized RL (MPR-RL)---that leverages prior experience collected on a subset of the problems in the MDP family to efficiently learn a policy on a new, previously unseen problem from the same family.
	Our approach learns prior distributions over the specific skill for each task and composes a family of skill priors to guide learning the policy in a new environment. %by comparing the similarity between the new task and the old ones. 
	%The RL policy generates a latent space representing the skill embedding that can be further decoded into real robot command sequences. 
	%We regularize the RL objective with relative Shannon entropy term based on the learned skill priors.
	We have evaluated our method on contact-rich peg-in-hole tasks shown in Figure~\ref{fig:franka}. %We show that our MPR-RL method can guide the policy learning over similar problems and that the composition of multiple skill priors accelerates training the RL policy. %Also, the incorporation of variable impedance~\cite{bogdanovic2020learning} guarantees the direct deployment of our approach on the real robot without further simulation-to-reality transfer.
	%\todo{Explain what makes this work relevant}
	\iffalse
	\begin{figure}[t!]
		\centering
		\subfigure[] {
			\includegraphics[width = 0.46\linewidth, height=3.4cm]{fig/maze.png}
			\label{fig:maze}
		}
		\subfigure[] {
			\includegraphics[width = 0.46\linewidth, height=3.4cm]{fig/franka_panda.png}
			\label{fig:franka}
		}
		
		\caption{An illustration of the experimental setup for two robot systems: \subref{fig:maze} 2D maze navigation task and \subref{fig:franka} a 7-DOF redundant Franka Panda for peg-in-hole tasks. We train multiple skill priors for a family of problems from demonstrated trajectories for each task.}
		\label{fig:experiment setups}
		\vspace{-0.4cm}
	\end{figure}
	\fi
	
	%%%%%%%%%%%%%%%%%%%%%%%%%%%%%%%%%%%%%%%%%%%%%%%%%%%%%%%%%%%%%%%%%%%%%%%%%%%%%%%%
	\section{\uppercase{Approach}}
	Our approach to transferring knowledge for RL is based on exploiting prior knowledge from demonstrations for learning a policy in a new task. The process is composed of two distinct phases: a prior learning phase and a task learning phase. In the task learning phase, we guide the policy learning to initially follow skill priors learned in the prior learning phase. For this, we regularize the RL objective with a relative entropy term based on the learned skill priors. %Below we formularize the problem setting, then we detail our skill prior learning, derive our multi-prior regularization method and finally propose a method to determine skill prior contribution during the task learning phase.

	We consider a family of tasks each formalized as a Markov decision process (MDP) defined by a tuple $(\mathcal{S}, \mathcal{A}, \mathcal{T}, r, \rho, \gamma)$ of states, actions, transition probability, reward, initial state distribution, and discount factor. A family of MDPs, $\mathcal{M}$, share the same state space and action space, while the dynamics and transition probabilities are different.
	
	We assume access to a dataset $D$ of demonstrated trajectories $\tau_i=\{(s_0,a_0),...,(s_{T_i}, a_{T_i})\}$ for each robotic task. We aim to leverage these trajectories to learn a skill prior $p_i(a_t|s_t)$ for each specific MDP $M_i$. Our objective is then to learn a policy $\pi_{\theta}(a|s)$ with parameter $\theta$ that maximizes the sum of rewards $G(\theta)$ for a new MDP $M_{new}$ by leveraging the prior experience contained in the dataset $D$.
	
	\iffalse	
	\subsection{Skill Prior Learning}
	\label{sec:skill latent space}
	In the prior learning phase, we use demonstrated trajectories $\tau_i$ to learn distributions over skill priors.
	We use a variational autoencoder (VAE)~\cite{Kingma2014AutoEncodingVB} model to learn a low-dimensional skill latent space $\mathcal{Z}$ from a dataset of pre-collected trajectories. The VAE model consists of a skill encoder $q(z|\boldsymbol{a})$ that outputs the latent representation $z$ of a skill and a decoder $p_{dec}(\boldsymbol{a}|z)$ that predicts a sequence of low-level actions $\boldsymbol{a}=\{a_t,\dotsb,a_{t+H-1}\}$ that the skill embedding $z$ represents, where $H\in\mathbb{N^+}$ is the action horizon. As described in \cite{pertsch2020spirl}, the skill prior model $p_{\boldsymbol{a}}(z|s_t)$ is used to generate a prior distribution over the latent space $\mathcal{Z}$ based on the state $s_t$. This distribution serves as guidance for the policy to determine which skills are worth exploring. Following~\cite{rezende2014stochastic} we train the VAE by optimizing the evidence lower bound (ELBO):
	\begin{equation}
		\label{equation:ELBO}
		\log{p(\boldsymbol{a})} \ge \mathbb{E}_q [\log{p(\boldsymbol{a}|z)}-\beta\left(\log{q(z|\boldsymbol{a})}-\log{p(z)}\right)],
	\end{equation}
	where $\beta$ is a hyperparameter used to tune the regularization term.
	\fi
	
	In skill prior RL (SPiRL)~\cite{pertsch2020spirl}, the learned skill prior is leveraged to guide learning a high-level policy $\pi_\theta(z|s)$ by introducing an entropy term. They propose to replace the entropy term of Soft Actor-Critic (SAC)~\cite{haarnoja2018soft} with the negated KL divergence between the policy and the prior.
	Similarly, our method uses the embedding space $\mathcal{Z}$.

	Using only one skill prior limits the method to policy learning in the same task as where the skill prior was learned. For this reason, we extend this approach from one learned skill prior to several skill priors learned in different tasks.
	To this end, we regularize the RL objective with a weighted sum of relative entropies:
		\begin{equation}
		\label{eq:sac objective}
		J(\theta)=\underset{\tau \sim \pi_\theta}{\mathbb{E}}\left[\sum_{t=0}^{T} \gamma^t r(s_t, a_t, s_{t'}) + \alpha \Gamma_t \right],\\
	\end{equation}
	where
	\begin{equation}
		\label{eq:multi-prior entropy}
		\Gamma_t = -\sum_{i=1}^{m}\omega_i D_{\text{KL}}(\pi_{\theta}(a_t| s_t),p_i(a_t|s_t)).\\
	\end{equation}
	$\omega_i$ is the adaptive weight and $\sum_{i=1}^{m}\omega_i = 1$. $p_i(a_t|s_t)$ are skill priors from different tasks of the family. This means that the policy is initially incentivized to explore according to a mixture of different skill priors depending on the weight factors. We learn the adaptive weights $\omega_i$ by training a discriminator for the most recent observed transition.  %While there are many options, in the next section we explain how we define the weight factors based on the current transition.

	\begin{figure}[t]
		\centering
		\subfigure[] {
			\includegraphics[width = 0.465\linewidth]{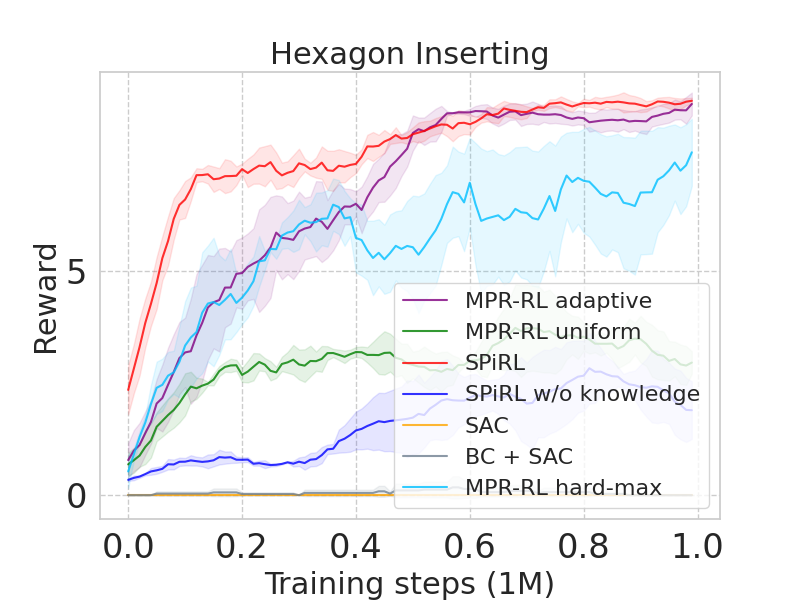}
			\label{fig:reward_hexagon}
		}
		\subfigure[] {
			\includegraphics[width = 0.465\linewidth]{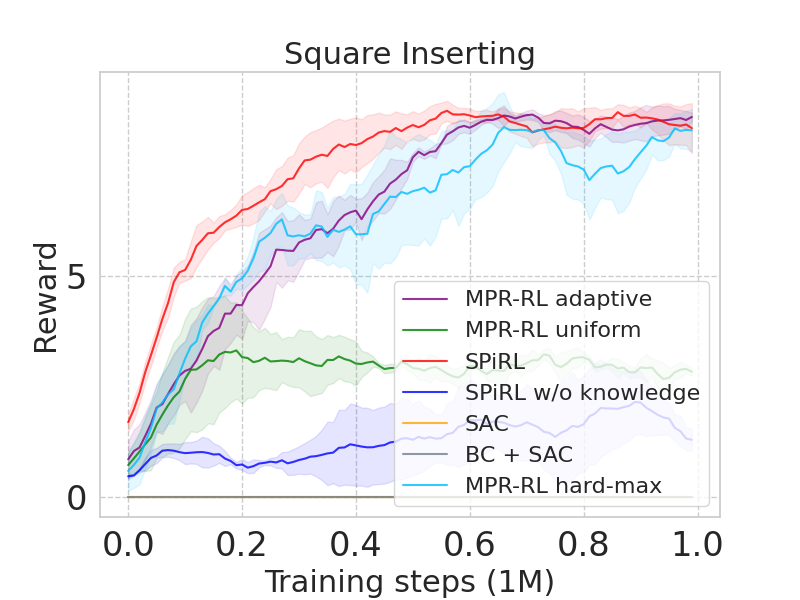}
			\label{fig:reward_square}
		}
		
		\caption{Comparing learning curves of MPR-RL with baseline methods: \subref{fig:reward_hexagon} rewards for inserting hexagon peg experiments; and \subref{fig:reward_square} rewards for inserting square peg experiments. Our MPR-RL method converges to good performance when used with adaptive prior weighting. Without prior knowledge from the target task, SPiRL fails to learn a solution policy. SAC from scratch and BC + SAC cannot finish the task. Shaded areas show standard deviation for three seeds.}
		\label{fig:rewards}
		\vspace{-0.2cm}
	\end{figure}

	\section{\uppercase{Results}}

	We evaluate the performance of MPR-RL using adaptive weights and compare with several baseline methods: (1) MPR-RL hard-max weight, where only the skill prior with the maximum task likelihood under the transition is used, (2) MPR-RL uniform weight, (3) SPiRL without prior knowledge from the target MDP, (4) Soft Actor-Critic (SAC), and (5) Behavioral Cloning with SAC (BC+SAC). When SPiRL has full access to the demonstration data from the target task, it is a baseline that is provided as a best-case \textit{oracle} target. % to try to reach. %We compare all methods' access to the demonstration data from the target or similar task in Table~\ref{tab:methods}.

	In the contact-rich peg-in-hole tasks, we use four different shapes as a family of MDPs: circular, hexagon, square and triangular as shown in Fig.~\ref{fig:franka}. The different shapes induce different contact dynamics and thus modify the transition probabilities of the MDPs in the family. We test two cases: (1) inserting hexagon peg by using prior knowledge from circular, square and triangular ones; (2) inserting square peg by using prior knowledge from circular, hexagon and triangular ones. These two cases offer the possibility to interpolate between some of the other shapes.
	
	%\subsection{Variable Impedance Action Space}
	%\label{sec:variable impedance action space}
	%To implement contact-rich tasks, we use a Cartesian impedance controller. 
	%In our experiments we use Pytorch~\cite{paszke2019pytorch} for RL training and the Robot Operating System (ROS) for communication between the robot controller and the RL agent. 
	%For a robot with $k$ joints, the observation vector $s_t$ is composed of: (1) joint positions $\boldsymbol{q}\in\mathbb{R}^k$ and joint velocities $\dot{\boldsymbol{q}}\in\mathbb{R}^k$, (2) end-effector position offset $\boldsymbol{e}\in\mathbb{R}^3$ and rotation $\theta_z$ in the $z$ direction, and (3) the environment contact force along the $z$ direction $F_{ext}\in\mathbb{R}$.	
	To train skill prior RL on the real robot directly, the system stiffness term $\boldsymbol{K}\in\mathbb{R}^{6\times6}$ is incorporated into the agent action~\cite{yang2022variable}. Therefore, we extend the policy action as the combination of end-effector pose $\boldsymbol{\xi}\in SE(3)$ in Cartesian space and variable stiffness matrix $\boldsymbol{K}$. Stiffness matrix $\boldsymbol{K}$ contains 6-dimensional end-effector stiffness coefficients. One extra null-space stiffness coefficient for the redundant robot is set as a constant value. 
	%We ignore rotation around the $x$ and $y$ axes of the end-effector frame and the corresponding variable stiffness components.
	Our 8-dimensional action space is thus composed of: (1) end-effector translations $\boldsymbol{x}\in \mathbb{R}^{3}$ in Cartesian space, (2) rotational angle $\theta_z \in \mathbb{R}$ around the $z$ axis, and (3) the diagonal coefficients $\boldsymbol{k}\in \mathbb{R}^{4}$ that determine the variable stiffness matrix $\boldsymbol{K}$ for the corresponding four Cartesian components.

	Fig.~\ref{fig:reward_hexagon} and Fig.~\ref{fig:reward_square} show the learning curves of our MPR-RL method and baselines over the real world hexagon and square peg-in-hole tasks respectively. Each method is trained using three different seeds. We can see that SPiRL with prior knowledge of the current MDP learns slightly faster, which is expected. MPR-RL policy with adaptive weight is able to learn to insert the target peg in the new MDP although it converges the reward plateau slower. MPR-RL with hard-max weight also succeeds in learning the policy, but shows fluctuating performance. The other methods fail to learn the skill. It demonstrates that our approach can transfer knowledge from a family of MDPs to a novel MDP successfully. In both experiments we conclude that the adaptive weight among a family of MDPs is essential for transferring knowledge for our MPR-RL policy.

\begin{figure}[t]
		\centering
		\subfigure[] {
			\includegraphics[width = 0.42\linewidth,]{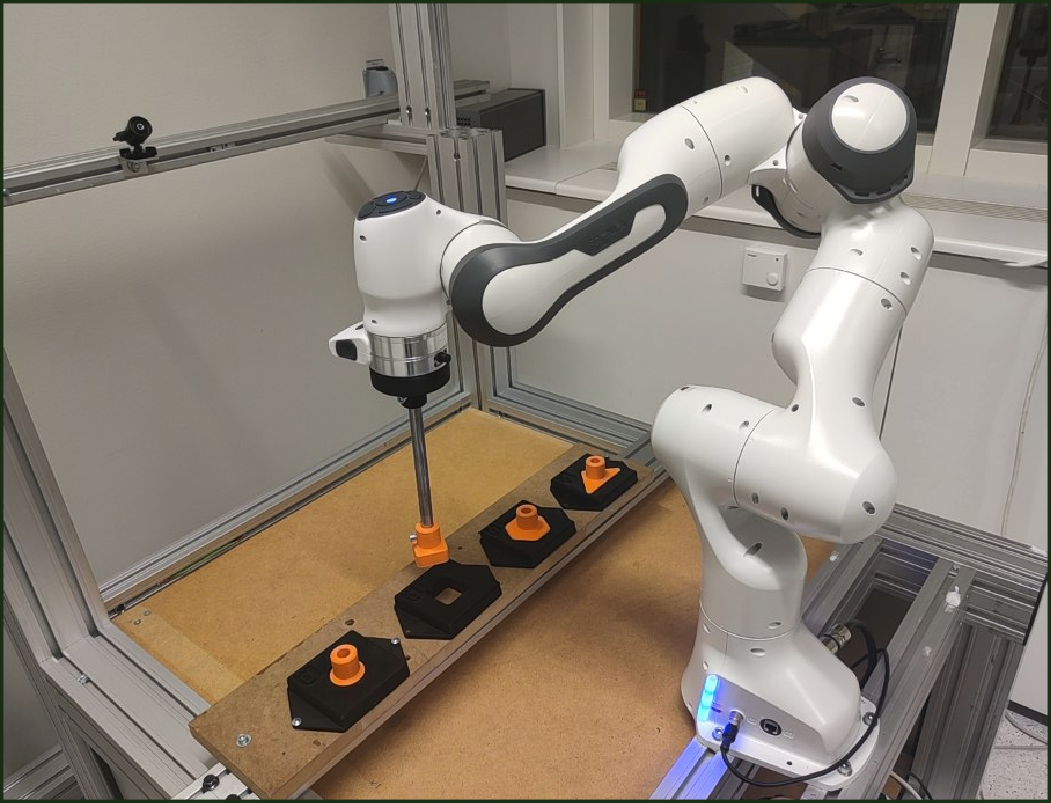}
			\label{fig:franka}
		}
		\subfigure[] {
			\includegraphics[width = 0.465\linewidth]{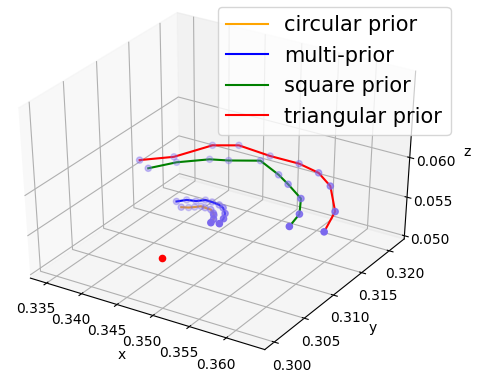}
			\label{fig:prior_traj}
		}

		\caption{\subref{fig:franka} Peg-in-hole tasks using Franka Panda arm; \subref{fig:prior_traj} We generate trajectories of 10 waypoints using prior knowledge from different MDPs.}
		\label{fig:real experiment}
		\vspace{-0.5cm}
	\end{figure}

	We compare one example of action sequences generated when using different skill priors in a hexagon peg-in-hole task shown in Fig.~\ref{fig:prior_traj}. Each trajectory is an action segment of 10 waypoints and the red point shows the position of the target hole. It can be seen that our MPR-RL method combines the knowledge from three different MDPs circular, square and triangular and transfer the inserting skill to a new hexagon MDP. In this specific case, the action sequence from multi-prior is more similar to the circular one.
	
	%As described in Section~\ref{sec:skill latent space}, the latent action $z$ presents a sequence of low-level actions $\boldsymbol{a}=\{a_t,\dotsb,a_{t+H-1}\}$. We investigate the influence of the latent space dimension $|\mathcal{Z}|$.
	
	%%%%%%%%%%%%%%%%%%%%%%%%%%%%%%%%%%%%%%%%%%%%%%%%%%%%%%%%%%%%%%%%%%%%%%%%%%%%%%%%
	\iffalse
	\section{\uppercase{Conclusions and future work}}
	\label{sec:conclusion}
	%\todo{Again describe the approach presented in this paper}
	%\todo{Again mention the advantages and what is novel compared to previous approaches}
	%\todo{Mention the implementation and the successful outcome of the experiments}
	%\todo{Potentially discuss options for future work\\
	%	- Don’t be too critical on your own work\\
	%	- Don’t be too enthusiastic about what else could and maybe should have been done.}
	\fi
	We have presented an approach that learns multiple priors for a family of similar MDPs and compose these priors to guide the RL training of a policy on a new MDP. Our approach learns prior knowledge over specific skills for similar tasks. By incorporating variable impedance into RL actions, we also show that our MPR-RL can be deployed directly on the real robot.
	
	%%%%%%%%%%%%%%%%%%%%%%%%%%%%%%%%%%%%%%%%%%%%%%%%%%%%%%%%%%%%%%%%%%%%%%%%%%%%%%%%
	
	%%%%%%%%%%%%%%%%%%%%%%%%%%%%%%%%%%%%%%%%%%%%%%%%%%%%%%%%%%%%%%%%%%%%%%%%%%%%%%%%
	
	%%%%%%%%%%%%%%%%%%%%%%%%%%%%%%%%%%%%%%%%%%%%%%%%%%%%%%%%%%%%%%%%%%%%%%%%%%%%%%%%
	\bibliographystyle{template/IEEEtran}
	\bibliography{template/IEEEabrv,references}

\begin{thebibliography}{1}
\providecommand{\url}[1]{#1}
\csname url@rmstyle\endcsname
\providecommand{\newblock}{\relax}
\providecommand{\bibinfo}[2]{#2}
\providecommand\BIBentrySTDinterwordspacing{\spaceskip=0pt\relax}
\providecommand\BIBentryALTinterwordstretchfactor{4}
\providecommand\BIBentryALTinterwordspacing{\spaceskip=\fontdimen2\font plus
\BIBentryALTinterwordstretchfactor\fontdimen3\font minus
  \fontdimen4\font\relax}
\providecommand\BIBforeignlanguage[2]{{%
\expandafter\ifx\csname l@#1\endcsname\relax
\typeout{** WARNING: IEEEtran.bst: No hyphenation pattern has been}%
\typeout{** loaded for the language `#1'. Using the pattern for}%
\typeout{** the default language instead.}%
\else
\language=\csname l@#1\endcsname
\fi
#2}}

\bibitem{peng2018sim}
X.~B. Peng, M.~Andrychowicz, W.~Zaremba, and P.~Abbeel, ``Sim-to-real transfer
  of robotic control with dynamics randomization,'' in \emph{2018 IEEE
  international conference on robotics and automation (ICRA)}.\hskip 1em plus
  0.5em minus 0.4em\relax IEEE, 2018, pp. 3803--3810.

\bibitem{hazara2019transferring}
M.~Hazara and V.~Kyrki, ``Transferring generalizable motor primitives from
  simulation to real world,'' \emph{IEEE Robotics and Automation Letters},
  vol.~4, no.~2, pp. 2172--2179, 2019.

\bibitem{du2021auto}
Y.~Du, O.~Watkins, T.~Darrell, P.~Abbeel, and D.~Pathak, ``Auto-tuned
  sim-to-real transfer,'' in \emph{2021 IEEE International Conference on
  Robotics and Automation (ICRA)}.\hskip 1em plus 0.5em minus 0.4em\relax IEEE,
  2021, pp. 1290--1296.

\bibitem{yang2022mpr}
Q.~Yang, J.~A. Stork, and T.~Stoyanov, ``Mpr-rl: Multi-prior regularized
  reinforcement learning for knowledge transfer,'' \emph{IEEE Robotics and
  Automation Letters}, vol.~7, no.~3, pp. 7652--7659, 2022.

\bibitem{pertsch2020spirl}
K.~Pertsch, Y.~Lee, and J.~J. Lim, ``Accelerating reinforcement learning with
  learned skill priors,'' in \emph{Conference on Robot Learning (CoRL)}, 2020.

\bibitem{haarnoja2018soft}
T.~Haarnoja, A.~Zhou, P.~Abbeel, and S.~Levine, ``Soft actor-critic: Off-policy
  maximum entropy deep reinforcement learning with a stochastic actor,'' in
  \emph{International conference on machine learning}.\hskip 1em plus 0.5em
  minus 0.4em\relax PMLR, 2018, pp. 1861--1870.

\bibitem{yang2022variable}
Q.~Yang, A.~D{\"u}rr, E.~A. Topp, J.~A. Stork, and T.~Stoyanov, ``Variable
  impedance skill learning for contact-rich manipulation,'' \emph{IEEE Robotics
  and Automation Letters}, 2022.

\end{thebibliography}
\end{document}